\documentclass{article}

\usepackage{arxiv}
\usepackage{graphicx}
\usepackage[utf8]{inputenc} 
\usepackage[T1]{fontenc}    
\usepackage{hyperref}       
\usepackage{url}            
\usepackage{booktabs}       
\usepackage{amsfonts}       
\usepackage{nicefrac}       
\usepackage{microtype}      
\usepackage{xcolor}         
\usepackage{subcaption}
\usepackage{graphicx}
\usepackage{color}
\usepackage{multirow}
\usepackage{wrapfig}

\usepackage{graphicx}
\usepackage{doi}
\usepackage{float}

\usepackage{tikz}
\usepackage{comment}
\usepackage{amsmath,amssymb} 
\usepackage{color}
\usepackage{wrapfig}
\usepackage[font=small,skip=0pt]{caption}

\usepackage[algoruled, resetcount, linesnumbered]{algorithm2e}

\usepackage{color}

\title{SVGraph: Learning Semantic Graphs from Instructional Videos} 

\author{Madeline C.~Schiappa \\
	Center for Research in Computer Vision\\ 
    University of Central Florida\\ 
	\texttt{madelineschiappa@knights.ucf.edu} \\
	\And
	Yogesh S. Rawat\\
  Center for Research in Computer Vision\\ 
  University of Central Florida, \\ 
  \texttt{yogesh@crcv.ucf.edu} \\
}  

\date{\vspace{-5ex}}
\renewcommand{\headeright}{}
\renewcommand{\undertitle}{\vspace{-5ex}}
\renewcommand{\shorttitle}{SVGraph}

\hypersetup{
pdftitle={SVGraph},
pdfsubject={cs.CV},
pdfauthor={Madeline C.~Schiappa, Yogesh S.~Rawat},
pdfkeywords={multimodal learning, deep learning, interpretability, graph, video understanding},
}

\begin{document}

\maketitle

\begin{abstract}
In this work, we focus on generating graphical representations of noisy, instructional videos for video understanding. We propose a self-supervised, interpretable approach that does not require any annotations for graphical representations, which would be expensive and time consuming to collect. We attempt to overcome ``black box" learning limitations by presenting Semantic Video Graph or \textit{SVGraph}, a multi-modal approach that utilizes narrations for semantic interpretability of the learned graphs. \textit{SVGraph} 1) relies on the agreement between multiple modalities to learn a unified graphical structure with the help of \textit{cross-modal attention} and 2) assigns semantic interpretation with the help of \textit{Semantic-Assignment}, which captures the semantics from video narration. We perform experiments on multiple datasets and demonstrate the interpretability of \textit{SVGraph} in semantic graph learning.

\end{abstract}

\section{Introduction}
The internet today hosts millions of instructional videos which can be made useful when analyzed automatically. However, designing an approach that learns an interpretable representation of videos without annotation is one of the biggest challenges. Most of the existing research in video domain has focused on specific tasks such as action detection \cite{carreira2017quo,tran2018closer}, temporal activity detection \cite{ding2020weakly,kopuklu2020watch} and retrieval \cite{duan2020omnisourced,xie2018rethinking,kopuklu2020watch}. However, these models are trained on large datasets with annotations specific to the task. This leads to issues of annotation cost \cite{yang2017suggestive,cai2021revisiting}, presence of annotation bias \cite{chen2021understanding,rodrigues2018deep}, lack of domain generalization \cite{wang2021generalizing,hu2020strategies,Kim_2021_ICCV}, and lack of robustness \cite{hendrycks2019benchmarking,hendrycks2021many}. 

Learning without annotations is often done by self-supervised learning (SSL), where a model is pre-trained on large-scale datasets without the need of labels \cite{jing2019selfsupervisedsurvey,survey}. Pre-training is done using a learning objective which is derived from the training samples itself, for example predicting playback speed \cite{Yao_2020_CVPR,wangjianglie2020,jenni2020video,Benaim_2020_CVPR,9412071} or frame order \cite{Fernando_2017_CVPR,Xu_2019_CVPR,10.1007/978-3-319-46448-0_32,Lee_2017_ICCV}. 
Videos are challenging not only because of time, but also because of greater variability in points of view, quality of video and quality of content.  These variations make it difficult for models to learn patterns by just relying on visual cues. Using multiple modalities alleviates some of these challenges and has been found effective in downstream tasks related to visual question-answering, retrieval, and action recognition \cite{alayrac2020selfsupervised,radford2021learning,alwassel2020self,5075633,miech2020endtoend,luo2020univl,sun2019videobert,miech2019howto100m}. However, none of these existing approaches have shown interpretability or a semantic understanding of the video. 

In image domain, \textit{scene graphs} have been proposed to provide interpretable learning in the form of graphs for entities and their relations \cite{chen2019knowledge,myeong2012learning,woo2018linknet,yang2018graph}. The challenge with these approaches is they require in-depth annotations that describe different entities and their relations to each other. Because they are also in the image domain, they do not directly translate to video with the addition of the temporal dimension. 
Obtaining frame-by-frame annotations for videos is very challenging and requires extensive resources in terms of time, cost and computation. This inspired us to develop an approach which learns interpretable graphical structure from videos without the availability of such annotations while taking advantage of longer duration of video.

\begin{figure*}[t!]
    \includegraphics[width=.99\linewidth]{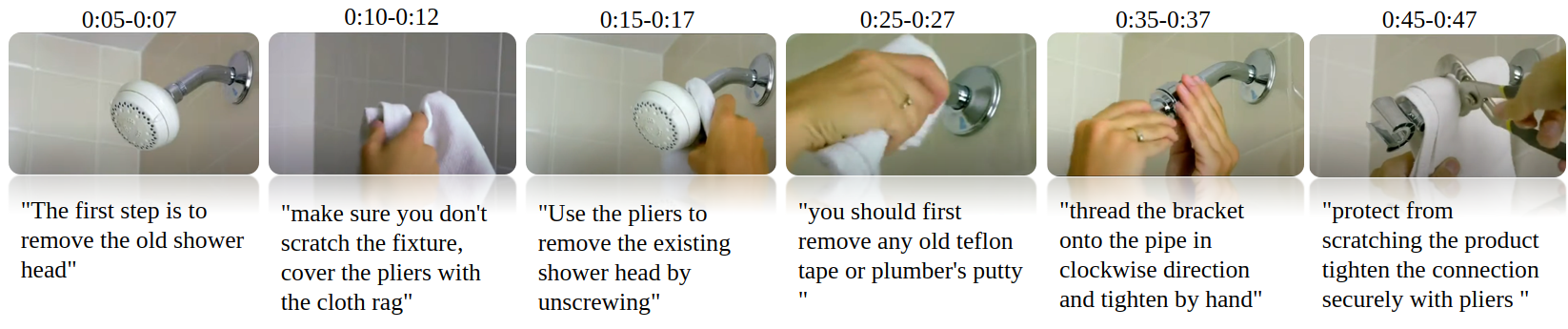}
    \caption{An example of the correlation between visual cues and audio cues in a video from an instructional video on ``How To change a shower head."}
    \label{fig:corr_example}
\end{figure*}

We propose \textit{SVGraph} which utilize multiple modalities to learn an interpretable representation. The multiple modalities will not always complement each other but sometimes they will also share information. For example, if a user is performing a task, the narration/audio will supplement that information (Figure \ref{fig:corr_example}). More specifically in the case of instructional videos, a user is more likely to narrate sub-activities when they appear on screen. This may indicate that visual stimuli at that time is important and should be attended to. Motivated by this, we utilize \textit{cross-modal attention} which facilitates one modality to attend the other. This enables enhancement of each modality with the help of other before their integration. 


Our proposed approach uses cross-modal embeddings and graph neural networks \cite{Gilmer} to model a video using multiple signals while generating an interpretable version of the learned representations. It learns joint-embeddings between visual, audio and textual signals obtained from automatic speech recognition (ASR) using cross-modal attention, without the need of any annotations. It uses multi-modal embeddings to build an interpretable graph with the help of convolutions for neighborhood-based message-passing and our \textit{Semantic Assignment}. We make the following contributions in this work,

\begin{itemize}
    \item We introduce \textit{SVGraph}, a novel approach for building a graphical representation of long instructional videos.
    
    \item SVGraph is trained using a self-supervised objective using multiple modalities without the need of annotations. 
    
    \item We propose a novel \textit{Semantic Assignment} mechanism which allows \textit{SVGraph's} graphical representation to be interpretable.

\end{itemize}

We perform our experiments on four different datasets and show both qualitative as well as quantitative evaluation validating the contributions of various components of \textit{SVGraph} and also demonstrate its effectiveness in semantic graph learning. 

\section{Related Works}
\subsection{Learning Complex Activities in Videos}
It is recognized that there are a set of sub-actions that often occur in the process of completing a long-complex action. Approaches in sub-action learning that are unsupervised often use deep clustering methods. Several approaches extend \cite{Caron_2018_ECCV} where cluster assignments acted as labels for images to sub-action detection in video. Each time segment from a video is assigned to a cluster and that cluster assignment acts as a sub-action label \cite{Sener2018,Kukleva2019,VidalMata2020}. One of the potential problems with these approaches is that the sub-action labels per video are not representing sub-actions at a global/dataset level, but rather at the local/video level. The authors in \cite{Kukleva2019} addressed this problem by assuming sub-activities will occur in the same temporal range for each video that represents the same long, complex activity. This would be concepts represented at the complex-activity level. Approaches that use global sub-action concepts with clustering can be computationally expensive. To reduce computational complexity, researches have proposed using a set of latent concepts to act as centroids and with each batch, a video or video segment is compared to the latent concepts to find the one(s) most similar \cite{Hussein2019,Caron2020}. While these approaches are able to capture both local and global information, they are not interpretable.

\subsection{Scene Graphs}
Scene graphs are methods that model the interactions between objects in a scene. They often use a pipeline of object detection for nodes, graph generation and then iterative updates of relationships between those nodes \cite{Dai_2017_CVPR,agarwal2020visual}. The most common approaches focus on refining the initial node embeddings extracted from object detection \cite{xu2017scene,Li2017SceneGG,gay2018visual,chen2019knowledge,li2018factorizable,gay2018visual}. These approaches are often highly spatial based \cite{Dai_2017_CVPR,xu2017scene} and the graphs are initialized as a fully-connected structure. None of these approaches address temporal relationships because of their focus on the image-domain, making it difficult to durectly apply to video. \cite{li2018factorizable,yang2018graph} have attempted to optimize graph generation by not assuming a fully-connected graph. This is useful because extending these image-based approaches to video would become even more challenging if the approach assumes a fully connected graph at initialization. All these architectures will learn object embeddings and relations embeddings through iterative updates, which we extend to video using SVGraph. Moreover, the requirement of dense annotations also differentiate these works from the proposed approach.

\subsection{Multi-Modal Representation Learning}
The most common tasks in multi-modal learning literature are representation learning and retrieval. Several works have employed a cross-modal learning objective to guide respective embeddings into a joint space \cite{survey}. One such approach is to predict whether a signal came from the same video \cite{arandjelovic2018objects,arandjelovic2017look,owens2018audio}, especially when using text as a modality \cite{Mithun2018,Radford2019,sun2019videobert}. These approaches use visual embeddings to predict whether a given text embedding from a set belongs to the respective video. Another is to use contrastive comparisons \cite{chen2020simple} between different modality embeddings, focusing on the joint-embedding space \cite{Peyre2020,NIPS2013_7cce53cf,sun2019learning,ma2019unpaired}. These aim to maximize the similarity between an embedding from one modality to the embedding of another. This approach is especially common in representation learning. We propose a new task for these methods, to learn an interpretable graph from the multi-modal embeddings.

\begin{figure*}[t!]
    \centering
    \includegraphics[width=.9\textwidth]{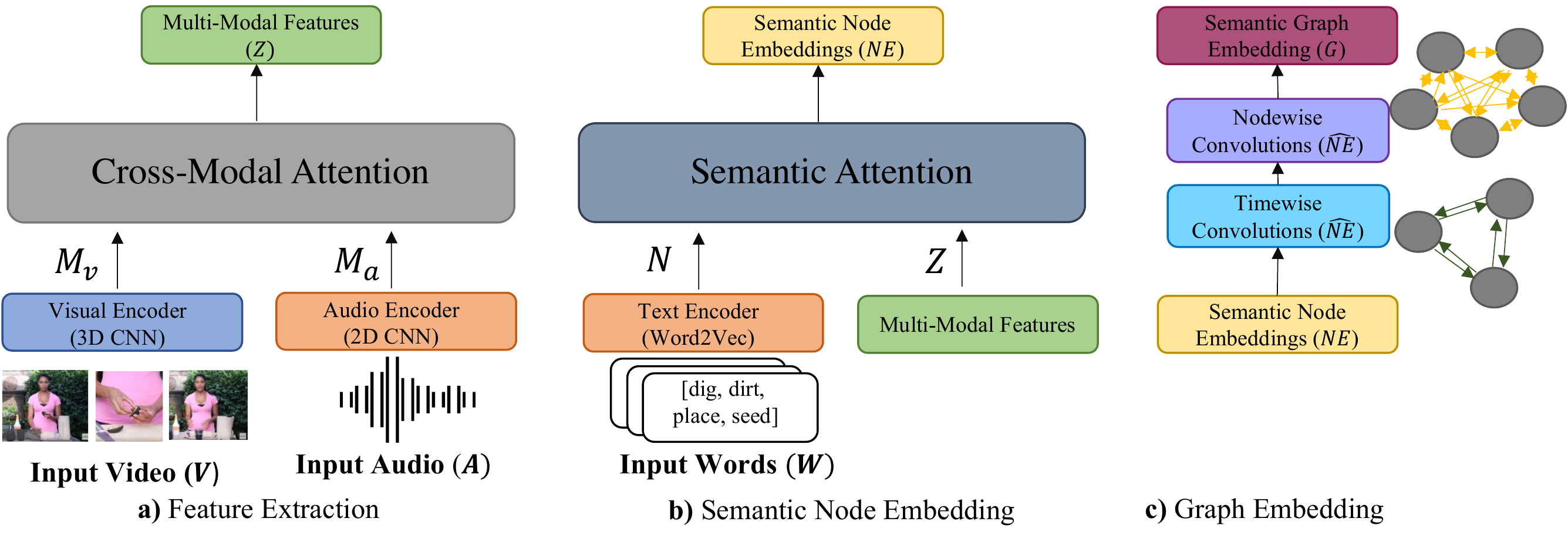}
    \caption{The proposed architecture for graph learning. Module (a) takes modality specific feature embeddings $M_v$ and $M_a$, and uses cross-modal attention to refine each modality's features then aggregating the refined features to generate a multi-modal feature embedding. Module (b) uses semantic node attention for node embedding between multi-modal features and and word features resulting in $\textit{NE}$. Module (c) uses neighborhood-based message passing resulting in refined $\hat{\textit{NE}}$.}
    \label{fig:architecture}
\end{figure*}

\section{Method}
Given a video $V \in \mathbb{R}^{T \times C \times H \times W}$ where $C$ are channels, $T$ is time and $H\times W$ are height and width of the frames, we want to learn a graphical representation $G$ that comprises of nodes $\hat{NE}$. Each video is first divided into multiple short clips with a sequence of frames which are encoded using a 3D CNN resulting in visual encodings $M_v$. Each video's aligned audio is extracted as Mel Spectrograms and encoded using a 2D CNN resulting in audio features $M_a$. 



Given an encoded video $M_v \in \mathbb{R}^{T \times C \times H \times W}$ and an encoded audio $M_a\in \mathbb{R}^{T \times C}$, we first learn a multi-modal embedding $Z \in \mathbb{R}^{T \times C}$. 
Next, words $W$ are encoded using Word2Vec from the narration in a video are used to initialize latent nodes $N$. Given a series of encoded words $N \in \mathbb{R}^{T \times N \times C}$, we attend to $N$ using $Z$. The semantic node embeddings $NE \in \mathbb{R}^{T \times N \times C}$ are further trained to learn the overall graph embedding, refining the semantic node embeddings $\hat{NE}$. Readout on the refined node embeddings $\hat{NE}$ is performed resulting in both a compact graph embedding $G$ and indices that map where the most relevant node embeddings were in the original set of nodes. Using \textit{Sementic Assignment}, these indices are used to map the original words $W$ to the maximally relevant node embeddings $\hat{NE}$. Using the graph embedding for each video, we use a triplet loss between the original video graph embedding $G_i$, an augmented version $\hat{G}_i$ and a randomly selected video $G_j$ to ensure consistency in learning. To generate $\hat{G}_i$, each input video $V$ is augmented and encoded to generate $\hat{M}_v$.
 
An overview of the proposed approach is shown in Figure \ref{fig:architecture}. Next we go through the details of the attention learning mechanism for multi-modal feature embedding in Section \ref{section:audio_visual} and semantic node embedding in Section \ref{section:node_embedding}. Then we go through our semantic node embedding refinement via message-passing in Section \ref{sec:message_passing}. Finally we go through the mapping of $\hat{NE}$ to $W$ via \textit{semantic assignment} in Section \ref{alg:semantic_backprop}.

\subsection{Cross-Modal Learning}
\label{section:audio_visual}
\begin{figure}[t!]
    \centering
    \includegraphics[width=.65\linewidth]{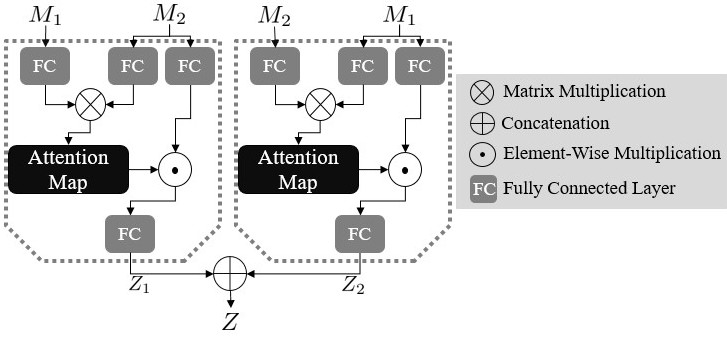}
    \caption{The cross-modal attention mechanisms used are a two-branch system where one modality $M_1$ is attended to by the other $M_2$ and vice versa. This results in a refined feature for both modalities $Z_1$ and $Z_2$ that when aggregated results in multimodal features $Z$.}
    \label{fig:crossmodal_attention}
\end{figure}

SVGraph uses multiple modalities to encourage learning to be focused on the most relevant activities in a video. For example, when the person is about to perform an important activity, they often narrate immediately prior to the activity or during the activity. The reverse may also be true where audio is more relevant to the activity when it is accompanied by a rapid change in motion of visible objects rather than when a person is standing still introducing themselves. We therefore utilize attention mechanisms for our multi-modal embeddings, attending to one modality via the other. This module is shown in Figure \ref{fig:crossmodal_attention} and is a two-branch attention mechanism where each branch uses one modality $M_1$ to attend to the other $M_2$. Each branch's procedure is similiar to self-attention in \cite{vaswani2017attention} but with the focus of cross-attention between two signals.

Input for this module are the modalities' feature representations extracted from their respective encoders, either a 3D CNN or a 2D CNN. These features comprise of $T$ time segments and a feature vector for each segment. We start by linearly projecting each modality in their respective branches. Then attention values $\alpha$ are used to correlated the embeddings which are calculated by taking the dot-product between $M_1$ and $M_2$ in each branch. Then $\alpha$ is multiplied by the linear projection of the modality we are attending to in the respective branch resulting in a refined feature embedding for each modality $Z_1$ and $Z_2$. This results in the original embeddings being refined where the points of greater similarity between the two modalities are emphasized. The output of each branch, $Z_1$ and $Z_2$, is then aggregated resulting in a final output which is a joint embedding of the two modalities $Z$.

\subsection{Semantic Attention}
\label{section:node_embedding}
In order to make our graph interpretable without annotations, we learn semantically relevant features for our nodes. We start by initializing nodes $N$ for our graph by using extracted features from the associated narration using Word2Vec. In order to refine and select the most relevant concepts from the narration, we attend to the semantic features using our multi-modal embeddings from Section \ref{section:audio_visual}. The extracted word features $N$ are associated with the same time segments as the other modalities and therefore their feature embedding will be further guided by the other modalities in a similar way to the multi-modal feature embedding. This approach is similar to a one branch attention mechanism as described in Section \ref{section:audio_visual} and shown in Figure \ref{fig:crossmodal_attention}. In this case, we are attending to our semantic nodes $N$ via the multi-modal features $Z$. This is done by calculating the dot-product between a linearly projected $N$ and linearly projected $Z$ to get attention values $\alpha$. These values measure the correlation between the semantic nodes extracted from narration and the multi-modal features where larger values indicate stronger similarity. We then attend to each node in $N$ with the attention values in $\alpha$ resulting in $\textit{NE}$. This results in the original embedding being modified where the nodes of greater similarity between the multi-modal embeddings are emphasized.   

\subsection{Message-Passing}
\label{sec:message_passing}
\begin{wrapfigure}[11]{l}{0.45\textwidth}
\vspace{-6mm}
\begin{algorithm}[H]
    \small
     \KwIn{$\textit{NE} \in \mathbb{R}^{T \times N \times C}$: Initialized Nodes}
     \KwOut{$I$, $\hat{\textit{NE}}$: Selection Indices, Nodes}
     
      \For{$l = 1$ to $L$}{
      Convolve Features: $\hat{\textit{NE}} = \text{CONV}(\textit{NE})$\\
      MP over $T$ \cite{hussein2019timeception}: $\hat{\textit{NE}} = \text{TimeCONV}(\hat{\textit{NE}})$  \\
      MP over $N$ \cite{Hussein2019}: $\hat{\textit{NE}} = \text{Node}\text{CONV}(\hat{\textit{NE}})$  \\
      MaxPool: $\hat{\textit{NE}}, I= \text{MaxPool}_{t\in T, n\in N}(\hat{\textit{NE}})$ \\
      }
     \Return $I$, $\hat{\textit{NE}}$\
     \caption{Message Passing}
     \label{alg:message_passing}
\end{algorithm}
\end{wrapfigure}
In order to learn relationships between the semantic node features $NE$, we utilize message-passing. As discussed in \cite{Gilmer}, convolutional layers can be used in a Message-Passing Neural Network (MPNN) framework to learn from graphs. Because we intend to only send messages between nodes within a respective neighborhood of space and time, using convolutions to update hidden states of each nodes is appropriate. We use iterative convolutional layers to allow message-passing between neighbouring nodes and time segments. This message-passing will allow the model to learn potential interactions between nodes and between a node over time, as represented by their features. It will also allow the model to learn the maximally relevant nodes and time segments. We use depthwise-convolutions, first proposed in \cite{chollet2017xception} and applied to graphs in \cite{Hussein2019}, to split the input and filter into groups, convolve each input with their respective filter and finally stack the convolved outputs together. This procedure is shown in Algorithm \ref{alg:message_passing}.

\subsection{Semantic Assignment}
 
\begin{figure}[t!]
    \centering
     \includegraphics[clip, width=.90\linewidth]{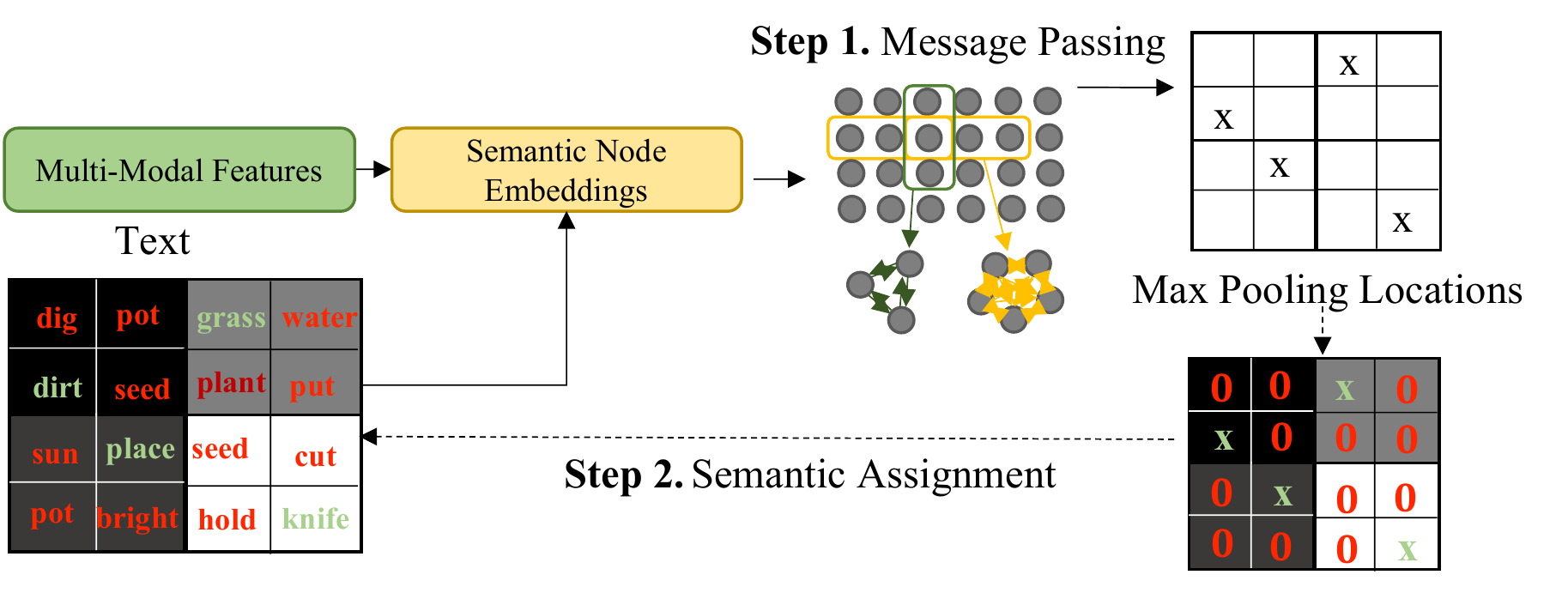}
    \caption{\textit{Semantic Assignment} starts with 1) semantic reverse mapping by collecting of max-pool indices after each message passing iteration and then Step 2) uses these indices to create a mask that is applied over the original set of words used to initialize nodes. The result is the final set of selected words that were embedded as nodes. }
    \label{fig:reverse_mapping}
\end{figure}

To make SVGraph interpretable, we propose \textit{Semantic Assignment}. During message-passing of the learned semantic node embeddings $n \in N$ with feature vectors $\textit{NE}_n \in \textit{NE}$, indexes $I$ refer to the maximally relevent nodes for the current instructional activity during the max pool operation. In order to represent interactions between objects, we interpret verbs/states as edges and their respective features as edge features. To do this, we must first map our selected nodes $N_I$ back to the original words so we can use their semantic meaning to assign nodes to entities or actions/states. An overview of this mechanism is shown in Figure \ref{fig:reverse_mapping}.
\begin{wrapfigure}[10]{l}{0.45\textwidth}
    \begin{algorithm}[H]
        \small
     \KwIn{$I, W$: Selection indices, words}
     \KwOut{$W^I$: Max relevant words}
     
      \For{$l = L$ to $1$}{ 
      Unpool: $I^{l-1} = \text{MaxUnpool}(I^l)$  \\
      Extract $W$: $W^l = [x_i \text{ for } i \in I^{l-1}]$\\
      }
      \Return  $W^I$\
      
      \caption{Semantic Assignment}
     \label{alg:semantic_backprop}
    \end{algorithm}
\end{wrapfigure} 
The extraction of $I$ is shown in Algorithm \ref{alg:message_passing} and the \textit{Semantic Assignment} is shown in Algorithm \ref{alg:semantic_backprop}. \textit{Semantic Assignment} uses the indices output from Algorithm \ref{alg:message_passing} to map backwards and retrieve the selected words $W_I$ from $W$ . Once we have the final set of words, we can use their respective features $\hat{\textit{NE}}$ for building graphs and to assign semantic meaning. It is important to use these features across time to measure node importance for each node, which can differ depending on a directed or undirected graph.

For directed graphs it is important to consider the difference between features over time $\hat{\textit{NE}} \in \mathbb{R}^{T \times N}$. To maintain time $t \in T$, we treat each occurrence of a word  at different time segments as separate nodes $n_t \in T$. Multiple occurrences of a word in one time segment are treated as the same by aggregating their respective activation values as: $\hat{\textit{NE}}_{n_t} =  \sum_{n \in N_t} \hat{\textit{NE}}_{n}$, 
where $N_t$ are all nodes that occur in time $t$. For undirected graphs, we want to compare nodes across all time segments. To do this, we calculate the average feature vector for each word over all occurrences as,
\begin{equation}
    \label{eq:undirected}
     \hat{\textit{NE}}_n = \frac{\sum_{t_\in T}\sum_{n \in N_t} \hat{\textit{NE}}_{n_t}}{\sum_{t_\in T}\sum_{n \in N_t} 1}.
\end{equation}
For both cases, the feature vector represents the activation of attention values for each node, which also indicates its relevance.

\subsection{Objective function}

In order to train the model, a readout procedure is performed on the graph to get an aggregate graph representation for the instructional video. This readout procedure takes the refined semantic node embeddings $\hat{NE}$ and aggregates them by a series of convolutions and max pooling operations to focus on the maximally relevant time segments and concepts. In order to train in a self-supervised fashion, we perform augmentation on the input video $M_v$ to generate $\hat{M_v}$. Using our framework we extract a graph embedding from readout $G_i$ using the original $M_v$ and $\hat{G_i}$ from the augmented version $\hat{M_v}$. We use a triplet loss \cite{schultz2003learning,triplet_loss} where $\hat{G_i}$ is the positive sample and $G_j$ is a randomly selected other video from the same batch as the negative sample. This learning objective aims to maximize the distance between the negative pair and minimize the distance between the positive pair while promoting discriminant feature learning. We used augmentations similar to \cite{chen2020simple} to generate a positive sample.

\section{Experiments}
\subsection{Experimental Setup}
For visual embeddings we use an I3D model initialized with weights pre-trained on ImageNet \cite{carreira2017quo}. For the audio embedding branch we use a 2D CNN initialized with weights pre-trained on acoustic scenery \cite{Koutini2019Receptive,KoutiniDCASE2019CNNVars}. We adapted augmentations from \cite{chen2020simple} for video to generate positive samples for the triplet loss during training. We trained models with stochastic gradient descent with a momentum of $0.9$, weight decay of $1e^{-7}$, and an cyclical learning rate \cite{smith2017cyclical} that used a base learning rate of $0.01$ and maximum learning rate of $0.1$. We trained for 50 epochs with an effective batch size of 128. 

\subsubsection{Datasets}
We perform our experiments on four different datasets.
\textbf{HowTo100M} \cite{Miech2019} is a large-scale dataset containing narrated instructional videos collected from YouTube. We chose a subset of videos that are under the activity category of `home and garden', `hobbies and crafts', and `computers and electronics'. Text is extracted from the ASR, or manually provided narration, downloaded from YouTube. In total, there are 19,662 videos used for training our approach. 
\textbf{COIN} \cite{COIN} is comprised of YouTube instructional videos. These videos have task labels, allowing us to make comparisons between different and similar tasks. The COIN dataset does not provide video narrations, therefore we used the video ids to retrieve ASR from YouTube. This resulted in 1,382 videos for training and 94 videos for testing. \textbf{YouCook2} \cite{ZhXuCoAAAI18} is a large, task-oriented, instructional video dataset for cooking. This dataset also does not provide video narrations. To collect the narrations, we again used the video ids to download ASR from YouTube for each video. This resulted in 662 videos for training and 238 for testing from 89 cooking recipes. \textbf{UCF101} \cite{soomro2012ucf101} is an action recognition dataset. These clips are short and do not have long-complex activities. These videos are annotated with action classes and therefore we use this dataset for our ablation experiments. We focused on videos that have audio signal available resulting in 4,839 videos for training and 1,944 videos for testing. 

\subsubsection{Metric}
\begin{wraptable}[8]{r}{0.4\textwidth}
\vspace{-4mm}
\caption{The overall average rouge-1 node overlap between within-task graphs and different tasks. The aim is for higher node overlap between same-task videos and lower for videos from different tasks.}
    \label{tab:node_overlap_table}
\resizebox{.4\textwidth}{!}{
    \begin{tabular}{c|c|c}
        \hline 
          & \multicolumn{2}{c}{Rouge-1 node overlap} \\
        \cline{2-3}
        Tasks  & Coin & YouCook2 \\ \hline
        Same Tasks & $86.67$  & $82.35$ \\
        Different Tasks & $35.87$ & $37.47$\\
        \hline
    \end{tabular}}
\end{wraptable}
To evaluate the quality of the learned graphs we measure the ability of our model to minimize the distance between videos from similar categories and maximize the distance between videos from different categories. To do this, we use a metric adopted from the natural language processing (NLP), the rouge-n metric. We use the rouge-1 \cite{lin2003automatic}, or the unigram overlap, between the nodes of a pair of videos.  
The complex activity recognition task on YouCook2 is evaluated using Precision@K \cite{ZhXuCoAAAI18} for K=5 and 10. Also, for activity recognition on UCF-101 we utilize accuracy scores. 

\begin{figure*}[t!]
    \centering
      \includegraphics[width=.42\textwidth]{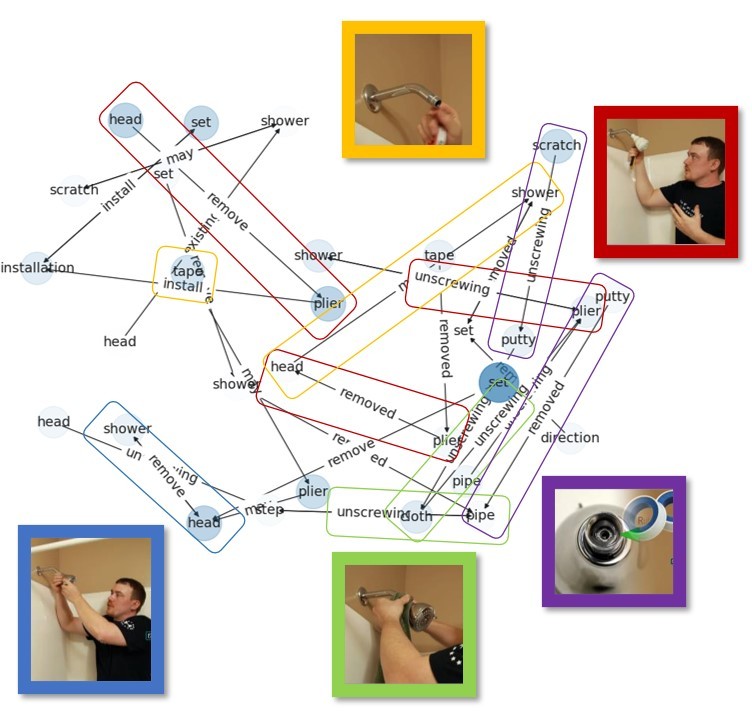}
      \qquad \qquad
      \includegraphics[width=.42\textwidth]{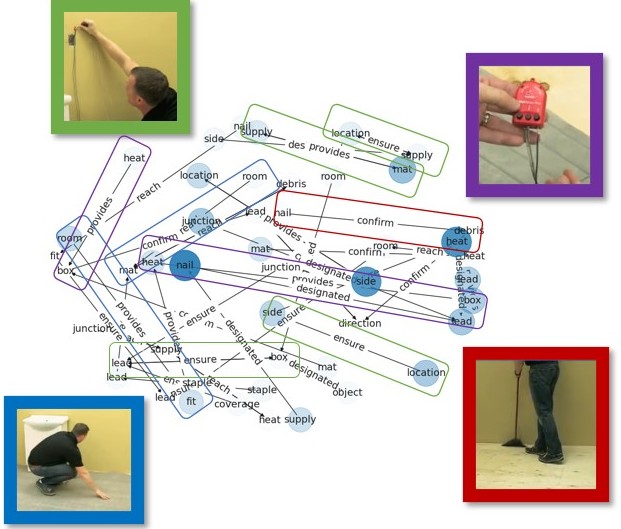}
    \caption{To the left, a visualization of ``how to install a shower head" and  to the right ``how to install a wood floor heating system". Each node's color illustrates its importance by weight where the darker the node, the greater the importance. Edges are determined by the similarity between a pair of objects and a node embedding for a word that is an action or state. The frames shown are extracted examples from the video and demonstrate some of the activities and relationships shown in the graph.}
    \label{fig:graph_visual}
\end{figure*}
\begin{figure}[t!]
    \centering
    \includegraphics[width=.85\linewidth]{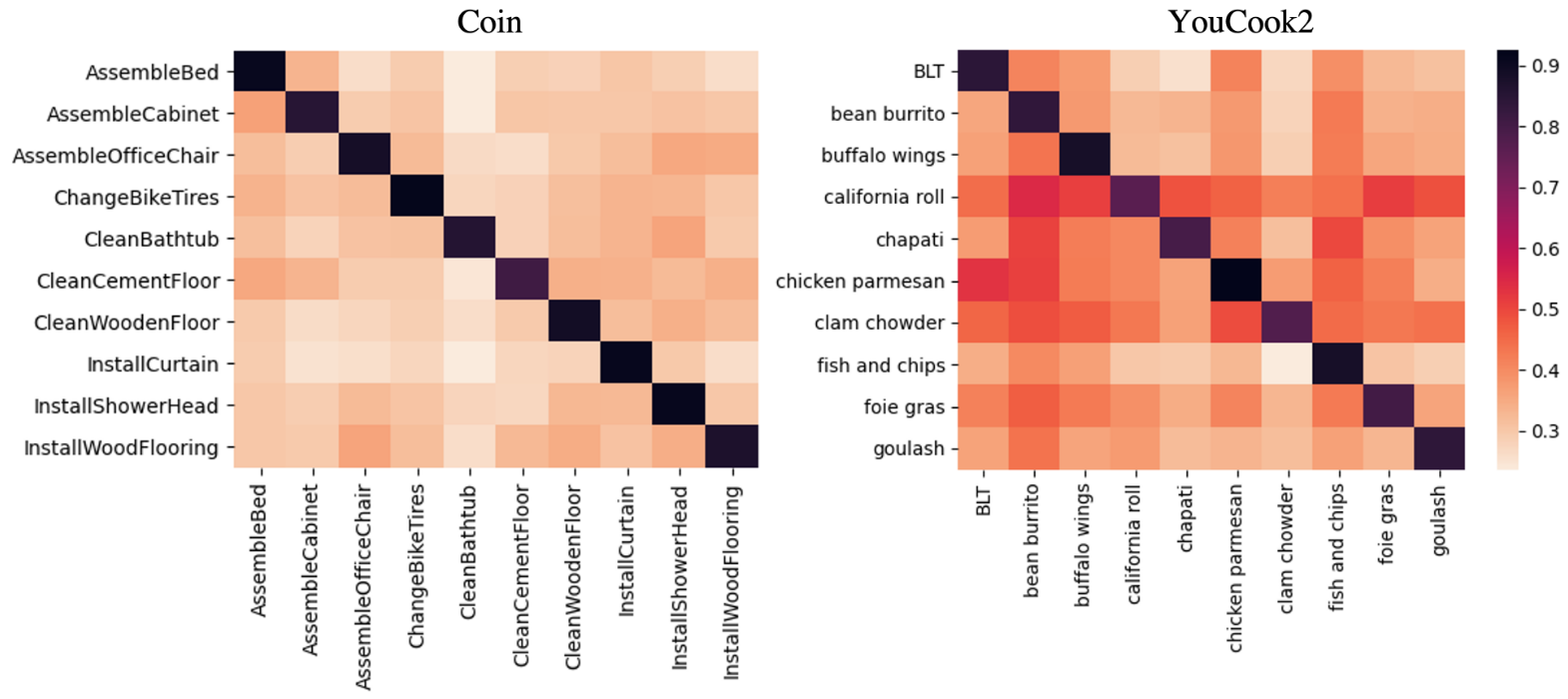}
    \caption{A visualization comparing videos from different tasks in both the YouCook2 dataset and the Coin dataset. The values shown are the average number of shared nodes between tasks. The darker the color, the greater the similarity. }
    \label{fig:node_overlap}
\end{figure}

\subsection{Graphical Analysis}
\begin{figure*}[t!]
    \centering
       \includegraphics[width=.42\textwidth]{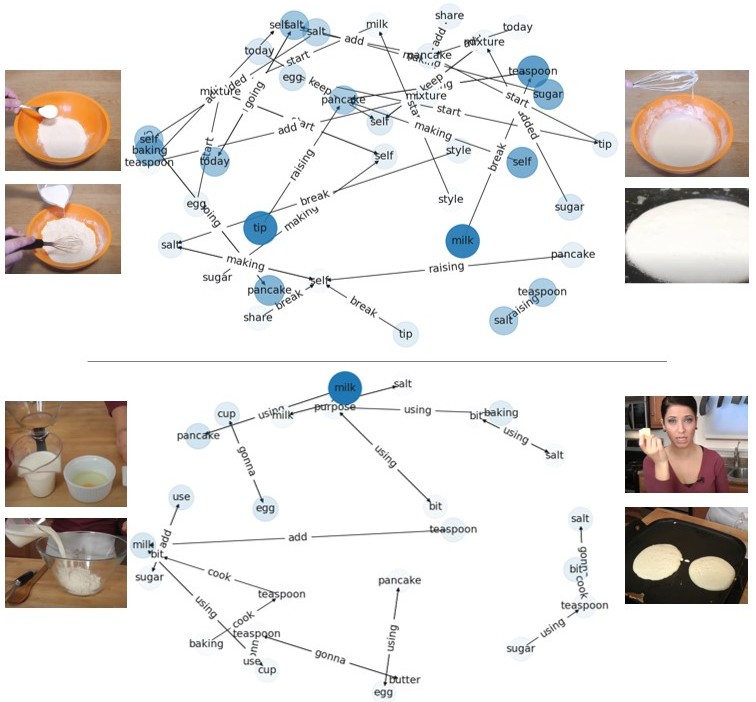}
       \qquad \qquad
       \includegraphics[width=.42\textwidth]{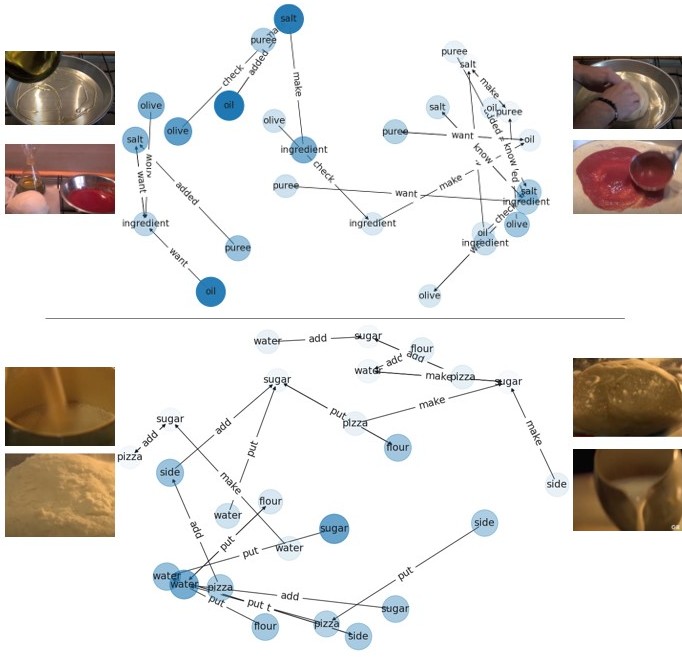}
    \caption{A visualization of learned graphs with corresponding sample frames for videos from same categories. (a) The two graphs on the top corresponds to two different videos about making pancakes and (b) the two graphs on the bottom corresponds to two different videos about making margarita pizza.}
    \label{fig:pancakes}
\end{figure*}

\subsubsection{Quantitative Analysis}
To make comparisons between graphs from videos of the same task or different tasks, we measure the node overlap between graphs. We use a rouge-1 \cite{lin2003automatic}, or the unigram overlap, between the nodes of one graph and the nodes of another. When there is complete overlap, rouge-1 would equal $1$ and where there is no overlap it would equal $0$. Figure \ref{fig:node_overlap} shows that the model is learning to pick maximally relevant nodes that are applicable to the specific task. While at some level there will be overlap based on words being unique to the task, the model could learn more generic terms that carry over all videos. For all tasks, the comparisons are shown in Table \ref{tab:node_overlap_table}. Both Figure \ref{fig:node_overlap} and Table \ref{tab:node_overlap_table} show that the model is learning to differentiate on key concepts between tasks. 

\begin{figure}[t!]
    \centering
     \includegraphics[width=.85\linewidth]{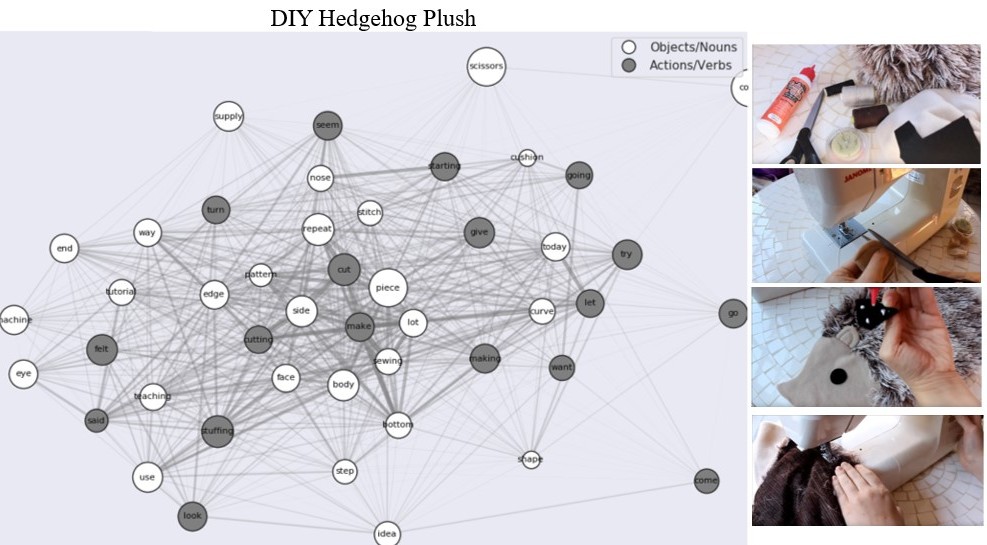}
    \caption{Undirected graphs for two instructional videos. These graphs aggregate node embeddings over time to generate an undirected graph. The larger the node size, the greater the importance for the overall activity. The thicker the edges, the stronger the relationships between two nodes are overall.
    }
    \label{fig:hedgehog}
\end{figure}

\begin{figure}
    \centering
    \includegraphics[width=.85\columnwidth]{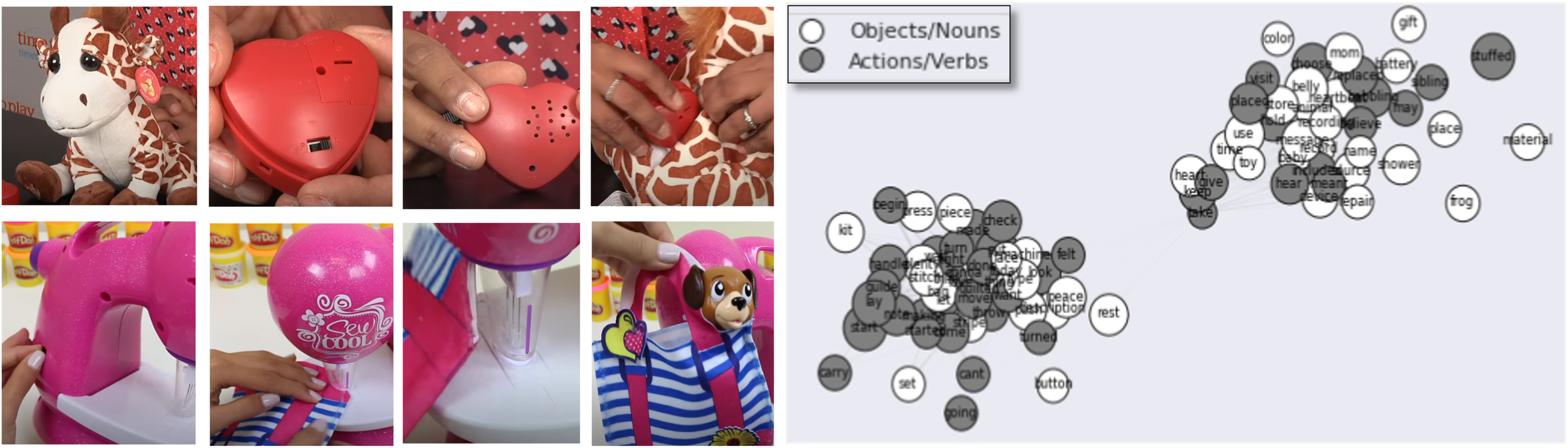} \\
    \includegraphics[width=.85\columnwidth]{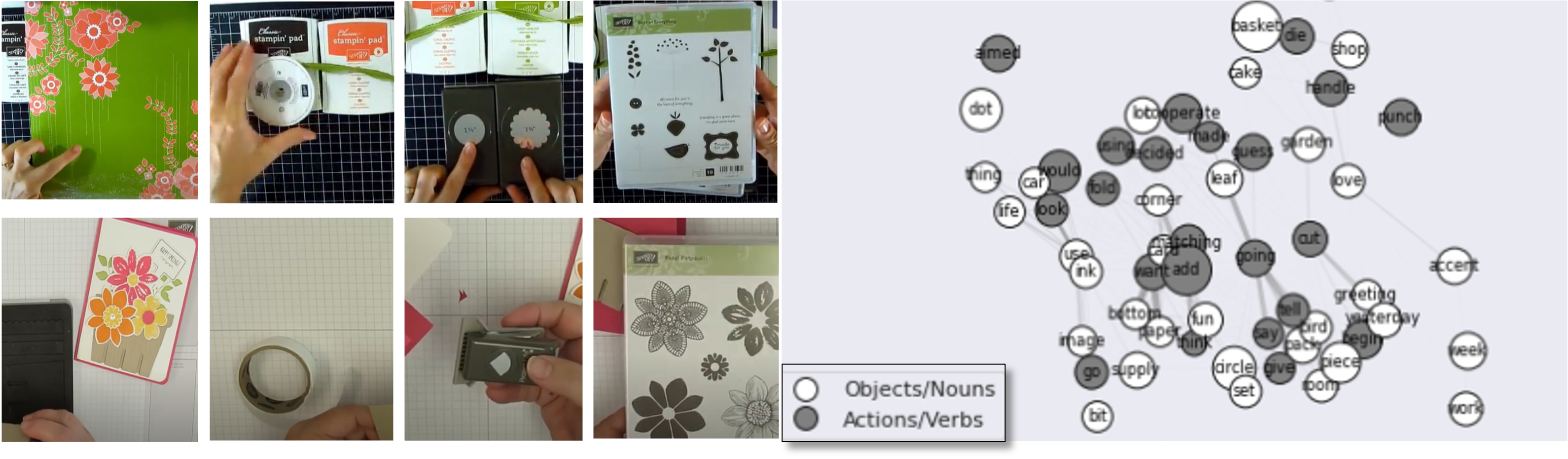}
    \caption{Aggregate undirected graphical representations of videos that either instruct the same task or different. Videos on the top row reflect a disjoint, two-clustered graphical representation while videos in the bottom row reflect an integrated graphical representation.}
    \label{fig:agg_complex_activites}
\end{figure}

\subsubsection{Qualitative Analysis}
Figure \ref{fig:graph_visual} shows example graphs learned for corresponding instructional videos. We observe that the key concepts appear in the video as important segments of instruction. For example in Figure \ref{fig:graph_visual}, the instructor in the video recommends using a cloth if the shower head is too tight to unscrew using a plier. The colors around sample frames are also around the nodes that illustrate that relationship. The most important concepts of the instruction appear to be repeated in the graph, demonstrating the model's ability to select nodes that are most relevant. Larger versions of these visualizations are available in the Appendix.
\begin{figure}
    \centering 
        \includegraphics[width=.80\textwidth]{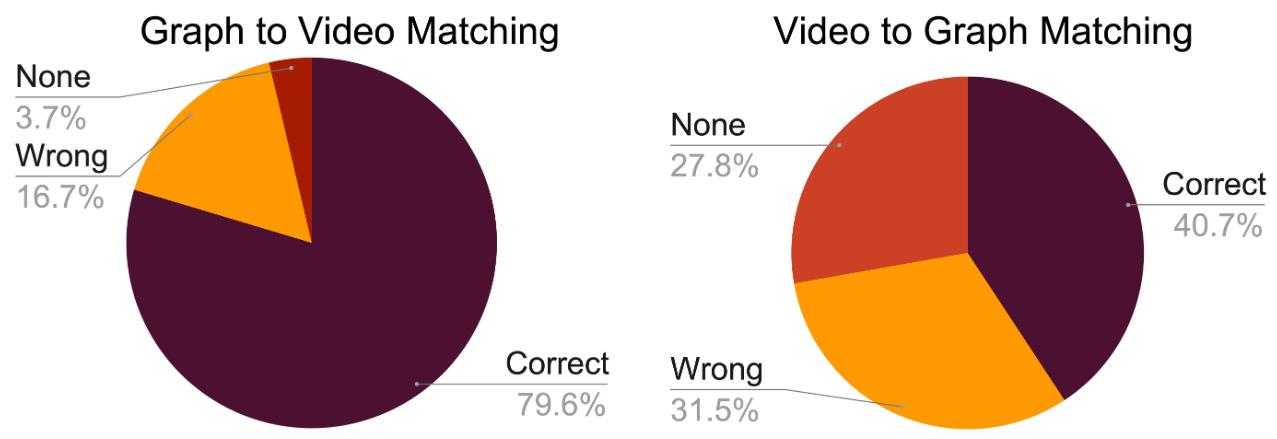}
    \caption{Results of our survey where (left) we ask users to identify the correct graph to video match and (right) correct video to graph match from multiple candidates.}
   \label{fig:survey_results}
\end{figure}
Instructional cooking videos in YouCook2 dataset have less variability in instructions as compared to other activities found in the Coin and HowTo100M dataset. Figures \ref{fig:pancakes} show visual comparisons between videos from YouCook2 on how to make margarita pizza and pancakes respectively. We observe that the graphs for making pizza on the right in  Figure \ref{fig:pancakes} have more content in common and the structure is more similar when compared with a different recipe such as `making pancakes' on the left in Figure \ref{fig:pancakes}. This supports that even through qualitative analysis, the graphs appear to be differentiating between tasks.

The proposed approach allows for both directed and undirected graphs. An example is an undirected graph in Figure \ref{fig:hedgehog} visualizing an unlabeled activity from the HowTo100M dataset. In this approach, the node features are aggregated over time. Therefore, nodes that are activated in more time segments have a greater importance to the overall activity. Here, we treat all words as nodes where edges are the strength of a relationship based on the cosine similarity. Node importance is visualized by the size of the node, where the greater the relevance the larger the node. The graph shown in Figure \ref{fig:hedgehog} visualizes the most important relationships towards the middle of the graph, also shown by the thickness of edges. Central to the graph, we see the activities and objects most commonly used such as “cut”, “cutting”, “sewing”. Less  common activities are more on the outer area of the graph such as “end”, “said”, “come”. 

This approach also allows for aggregation of graphical representations. Figure \ref{fig:agg_complex_activites} shows an aggregation of two graphs from videos instructing the same activity (top) and shows aggregation between videos instructing two different activities (bottom). In the same activity aggregation, you can see more connections and overlap between the node features while in the different activities there is absolutely no overlap or similarity between the nodes. This shows SVGraph is learning to distinguish between different activities at a global level. 

\subsubsection{User Study}
To further evaluate the quality of learned graphs, we performed an user study. We surveyed three different aspects of our work: 1) graph to video matching, 2) video to graph matching, and 3) graph quality. 
We first presented users with a graph and asked them to choose which of the given videos the graph best represented with an option that the graph represented none of the provided videos. We then did the reverse and presented them with a video and a choice of multiple graphs to match with an option of no graph. Finally, we presented graph and video pairs and asked users to rate how well the respective graph represented its video on a scale from 1-10. Video and graph samples were randomly selected from the Coin and Youcook2 dataset. For each question, four options were presented to the user. The results for the 30 participants are shown in Figure \ref{fig:survey_results}. Users were better able to match a graph to a video but struggled with the reverse. Overall, the correct match was made a majority of the time. When users were asked to rate the quality of randomly chosen graph-video pair from a scale of $1$ to $10$, the average rating was $6$.
  
\subsection{Ablations and Discussion}
\subsubsection{Self-Supervised Objective}

\begin{figure}[t!] 
\centering

\includegraphics[width=.85\linewidth]{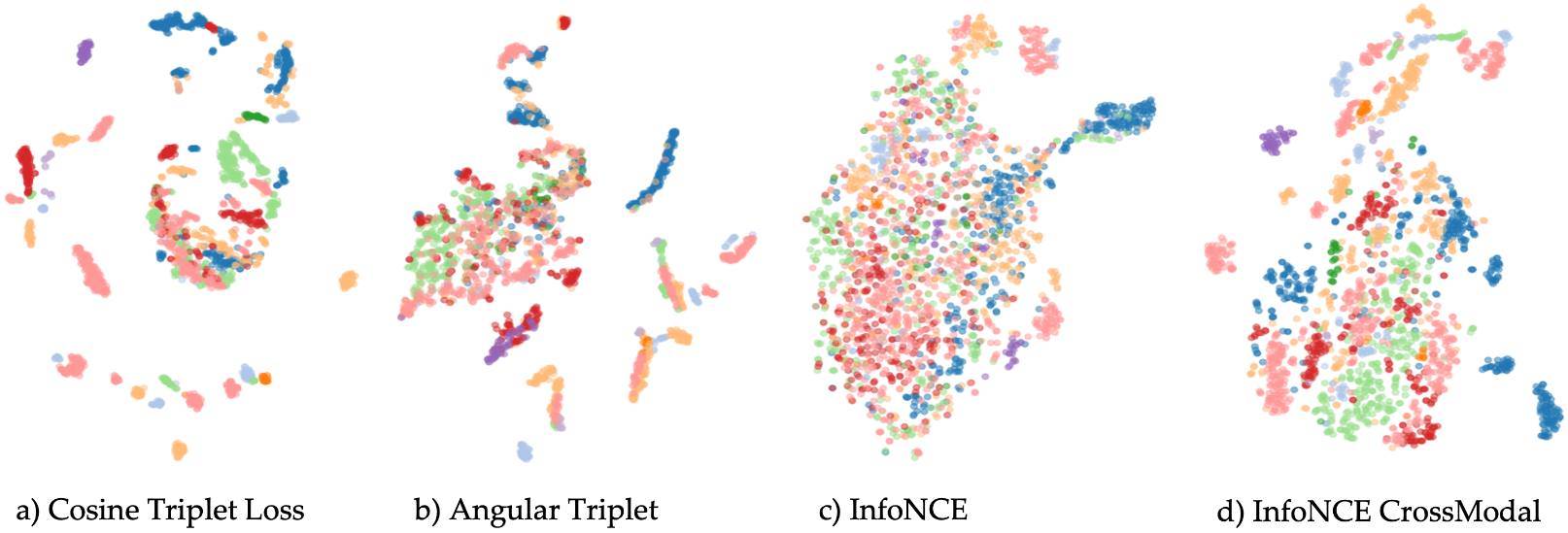}
\caption{A comparison between the features of videos from a subset of videos in the Coin dataset based on the objective function the model was trained on. Features were extracted from the joint-embedding of video and audio.}
\label{fig:losses}
\end{figure}

We experimented with multiple self-supervised loss functions to train our approach: cosine-triplet loss \cite{hermans2017defense}, angular-cosine triplet loss \cite{wang2018cosface}, and a noise-contrastive estimation (NCE) \cite{mnih2013learning}. Each loss uses an augmented version of the frames as a positive sample while all other samples in the batch as negative samples. We also tested a cross-modal NCE loss that treat the visual and audio branch separately when attending to text. The resulting joint embedding between audio+text $\text{NE}_a$ and video+text $\text{NE}_v$ are used as positive samples for the NCE loss. Figure \ref{fig:losses} shows a visualization of learned features for sub-activities in the COIN dataset extracted from each model. The cosine triplet loss shows the most distinct groupings of the activities while the cross-modal shows the most separation of different activities. Without the cross-modal positive samples, the NCE loss alone does not show a good separation of different tasks.

\begin{wraptable}[8]{r}{0.4\textwidth}
    \vspace{-4mm}
    \caption{Ablations for the use of audio and visual features.}
    \label{tab:supevised_ablations}
    \resizebox{.4\textwidth}{!}{\begin{tabular}{l|c|c}
    \hline
     & \multicolumn{2}{c}{Precision} \\ 
        Method & P@5 & P@10 \\ \hline
         SVGraph w/Video & $6.5\%$ & $12.6\%$\\
         SVGraph w/Audio  & $7.1\%$ & $13.7\%$ \\
         SVGraph w/Video+Audio & $9.1\%$ & $16.1\%$ \\
         \hline
    \end{tabular}}
\end{wraptable}

\subsubsection{Video and Audio Joint Embedding}
To better understand if using both audio and video will improve learning, we trained SVGraph up to the visual-audio joint embedding $\text{VA}$ on the YouCook2 dataset, predicting on the overall video activity. We used a clip length of $128$ and a fixed word length of $15$. We compare results to a video only prediction using I3D \cite{carreira2017quo}. The results in Table \ref{tab:supevised_ablations} indicate that using both audio and video improve performance compared to using one over the other.

\subsubsection{Cross-Modal Attention Mechanism}
We use UCF101 \cite{soomro2012ucf101} which is annotated with action classes for this ablation. We only used videos that contained audio, resulting in 4893 for training and 1944 for testing. Visual features and audio features are extracted in the same way as prior experiments. For efficiency, we reduce clip lengths to $t=16$ with a smaller batch size of $8$. A joint-embedding is learned through either multiplication, summation, concatenation, a one-branch attention focused on attending to video $M_v$ or our proposed \textit{cross-modal attention}. Because our \textit{cross-modal attention} combines the attended outputs of the two branches, we experimented with the different possible variations. The one-branch attention module was used where video features $M_v$ were the query and the audio features $M_v$ were the key. It was also used for self-attention on $M_v$ only. The resulting joint-embedding between visual and audio features is then passed through a FC layer to predict activity. 
The results in Table \ref{tab:joint_embed} further indicate that using both audio and video improves performance. The results also indicate that \textit{cross-modal attention} with concatenation performed best out of all the other methods described.

\begin{wraptable}[11]{r}{0.4\textwidth}
    \vspace{-5mm}
    \centering 
    \caption{Ablations on learning joint-embeddings.}
    \label{tab:joint_embed}
    \resizebox{.4\textwidth}{!}{\begin{tabular}{l|c}
        \hline
         Method & Accuracy  \\ \hline
         I3D \cite{carreira2017quo} &  $33.2\% $ \\ 
         Self-Attention \cite{vaswani2017attention} & $63.2\%$ \\
         \hline
         One-Branch Attn. & $21.1\%$ \\ 
         Sum & $29.1\%$ \\
         Multi &  $30.9\%$ \\
         Concat & $41.3\%$ \\
         \hline
         Cross-Modal Attn. + Sum & $68.2\%$ \\
         Cross-Modal Attn. + Multi & $75.9\%$ \\
         Cross-Modal Attn. + Concat & $\mathbf{78.7\%}$ \\
         \hline
    \end{tabular}}
\end{wraptable}
\subsubsection{Global vs. Local Representation}
To analyze how local or global the learned representations are between tasks, we combined the nodes and node embeddings of two videos that share the same instructional task according to their HowTo100M task description. Figure \ref{fig:agg_complex_activites} shows two aggregated graphical representations, each of two videos that are labelled to be performing the same task. The top row of Figure  \ref{fig:agg_complex_activites} shows videos described as how to make a stuffed giraffe, but on inspection, one is instructing users on a ``Design Deluxe Sewing Studio Playset" and the other is providing a review on ``My Baby's Heartbeat Bear (Giragge)". When observing their combined graphical representation, there are two distant clusters that formed, further supporting their difference. The bottom row of Figure \ref{fig:agg_complex_activites} shows videos described as how to make berry card and both videos were instructing that activity. When observing their combined graphical representation, there is a more intertwined graphical representation. While our approach is instance-based, there appears to be some level of global learning of complex activities based on the aggregate graphical representation, even more so when differentiating between activities.

\section{Conclusion}
We propose a new task, representing instructional videos in semantically meaningful graphical form without the use of annotations. To solve this problem, we propose SVGraph, a framework to address this challenge that uses multiple-modalities from visual, sound and text in video. We incorporated cross-modal attention to improve the learning of joint-embeddings between these modalities. We proposed a novel technique \textit{Semantic Assignment} to make these representations semantically interpretable. While it is a challenging problem, we demonstrate its feasibility which opens up an interesting research direction in video understanding. 

\bibliographystyle{splncs04}
\bibliography{mybibliography}

\clearpage

\appendix

In this supplementary we provide additional details about the model implementation in Section \ref{supp_model_details} and additional results in Section \ref{supp_additional_results}. These results are enlarged versions of several semantically interpreted graphs for easier viewing and understanding. 


\section{Model Details}
\label{supp_model_details}

\paragraph{Cross-Modal Attention Implementation Details}
In the case of video and audio as the two modalities, we start by extracting visual features $M_v$ using a pre-trained 3D CNN \cite{carreira2017quo}. We extract audio features from the same time segments as video $M_a$ using a CNN pre-trained on acoustic scenery \cite{Koutini2019Receptive}. To retain temporal information for both audio and video, we keep all remaining time segments rather than average pooling them before final output. For example, 256 frames would result in 32 time segments, which at an FPS of 30, is approximately 8 seconds.  Using cross-modal attention, we pass video features $M_v$ as a query and audio features $M_a$ as the key and value in one branch and $M_a$ as a query and $M_v$ as the key and value in the second branch. We then aggregate each branch output and project the features down to the original channel size, resulting in a set of multi-model features $Z$. 

\begin{align}
        & \mathbf{\alpha}_1 = \text{ReLU}\left((W^{Q_1} M_1) * (W^{K_1} M_2)\right)  \nonumber  \\
        & Z_1 = \text{ReLU}\left(W^{\text{out}_1}(\mathbf{\alpha}_1 \cdot W^{V_1}M_2)\right) \nonumber \\
        \nonumber \\
        & \mathbf{\alpha}_2 = \text{ReLU}\left((W^{Q_2} M_2) * (W^{K_2} M_1)\right) \label{eq:attend}   \\
        & Z_2 = \text{ReLU}\left(W^{\text{out}_2}(\mathbf{\alpha}_2 \cdot W^{V_2}M_1 \nonumber)\right) \\
        \nonumber \\
        & Z = Z_1 \oplus Z_2 \nonumber
\end{align}

\paragraph{Message Passing via Depthwise Convolutions}
We use depthwise-convolutions \cite{chollet2017xception,Hussein2019} to split the input and filter into groups, convolve each input with their respective filter and finally stack the convolved outputs together. It applies a 1D convolution with $C$ kernels of $k^T \in \mathbf{R}^{t \times 1 \times 1}$ to the temporal dimension then a 1D convolution with $C$ kernels of $k^N \in \mathbf{R}^{1 \times n \times 1}$. A max-pool operation is then carried over the resulting time dimension and node dimension respectively. This procedure is repeated $n$ times. After each iteration, we max-pool over nodes to select those that are maximally relevant. The output of these iterations $l \in L$ is a set of indices for each iteration $I^l \forall l \in L$ and updated semantic node embeddings $\hat{\textit{NE}} \in \mathbb{R}^{T \times N \times C}$.

\paragraph{Triplet Loss and Augmentation}
We use a triplet loss that uses an augmented version of a video $f_i$ as its positive sample $f^+_i$. A randomly selected sample from the same batch is used as a negative sample $f_i^-$. 

\begin{equation}
     L = \frac{1}{N} \sum^N_{i = 1} - \text{log} \frac{e^{f_i^T f^+}}{e^{f_i^T f^{+}} + e^{f_i^T f^{-}} } \label{eq:loss_func} 
\end{equation}

Equation \ref{eq:loss_func} aims to maximize the distance between the negative pair $(f_i, f_i^-)$ and minimize the distance between the positive pair $(f_i, f_i^+)$ using,
\begin{align}
    & {f_i^{T} f^{+}} =  \Vert f_i \Vert \Vert f_i^+ \Vert \text{cos}(\theta^+_i)-m \\ 
    & {f_i^{T} f^{-}} =  \Vert f_i \Vert \Vert f_i^- \Vert \text{cos}(\theta^-_i)
\end{align}

where $m$ forces $\text{cos}(\theta^+_i)-m > \text{cos}(\theta^-_i) $, promoting discriminant feature learning. For generating positive samples, we apply the same augmentations to all frames so that the entire video is consistent in color, brightness, and size. It starts with a random resize crop to a fix height and width. The cropped video is then augmented using color-jittering by randomly altering the brightness, contrast, saturation and hue. We then randomly choose, with a 50\% probability, whether to horizontally-flip the frames and whether to convert the frames to gray-scale. Finally a Gaussian blur is applied on the augmented sample.

\paragraph{Data Pre-Processing}
For each video, we uniformly sampled a segment of 256 frames at 15 fps. We aligned audio and text using the starting point and ending point in time for each video segment. We extract each word from narration and embed them using a pre-trained Word2Vec model \cite{mikolov2013efficient}. Word2Vec representations were chosen because the cosine similarity between vectors indicates the level of semantic similarity between their respective words. It was also chosen because the ASR captions are often lacking proper grammatical structure, which are often needed for more sophisticated pre-trained natural language processing (NLP) models. For extracted audio clips, we calculated the spectrogram and converted the frequencies to the mel scale using \texttt{torchaudio} from PyTorch  \cite{NEURIPS2019_9015}. 



\begin{figure*}[t!]
    \centering
    \includegraphics[width=.99\linewidth]{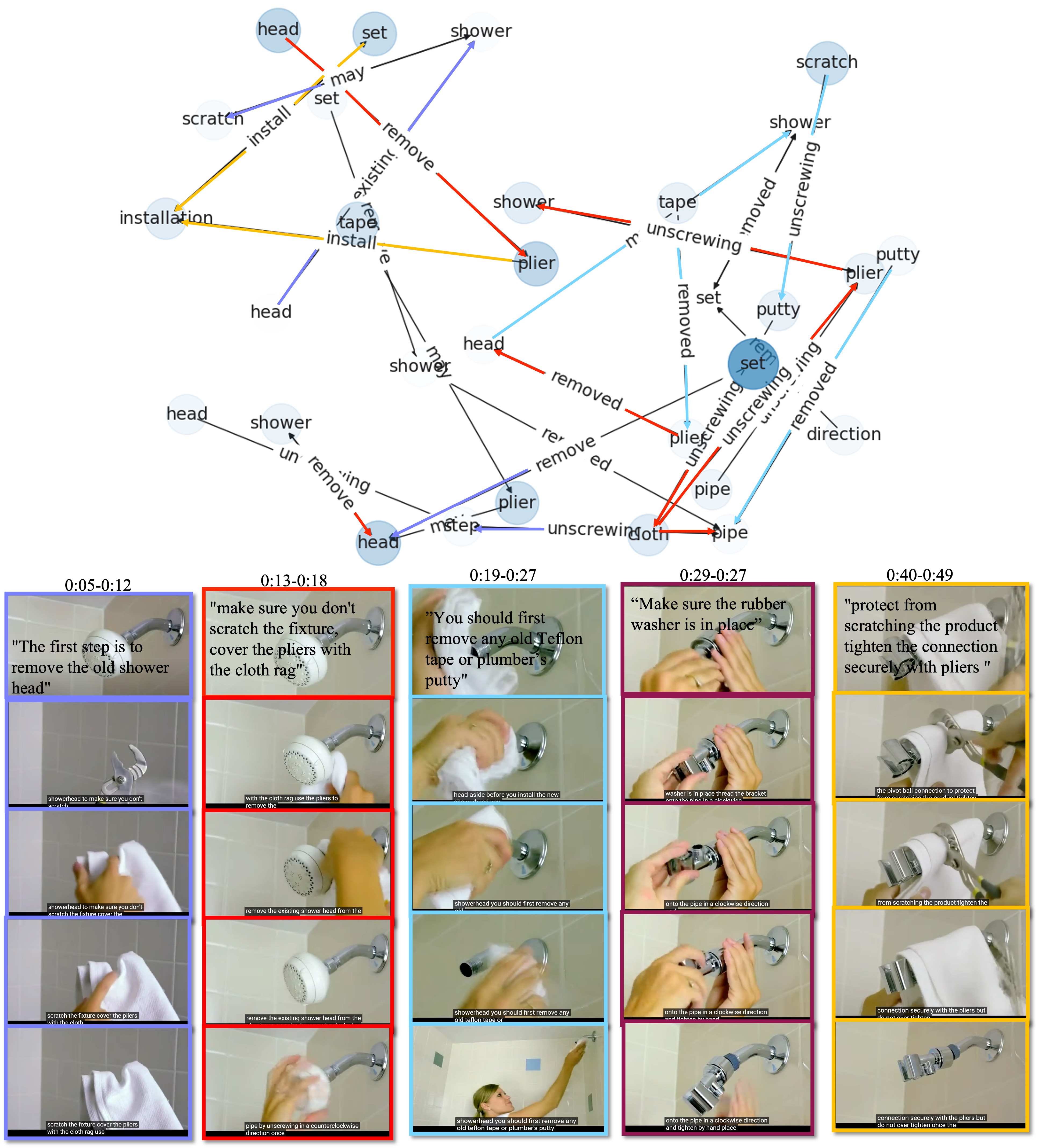}
    \caption{A larger example ``Change Shower Head" and its interpreted graph using SVGraph. The bottom row of images are sample frames from the video and associated narration. The instructions occurring at those times are highlighted by the same color to help guide understanding in this example.}
    \label{fig:Change_Shower_head}
\end{figure*}

\begin{figure*}
    \centering
    \includegraphics[width=.99\linewidth]{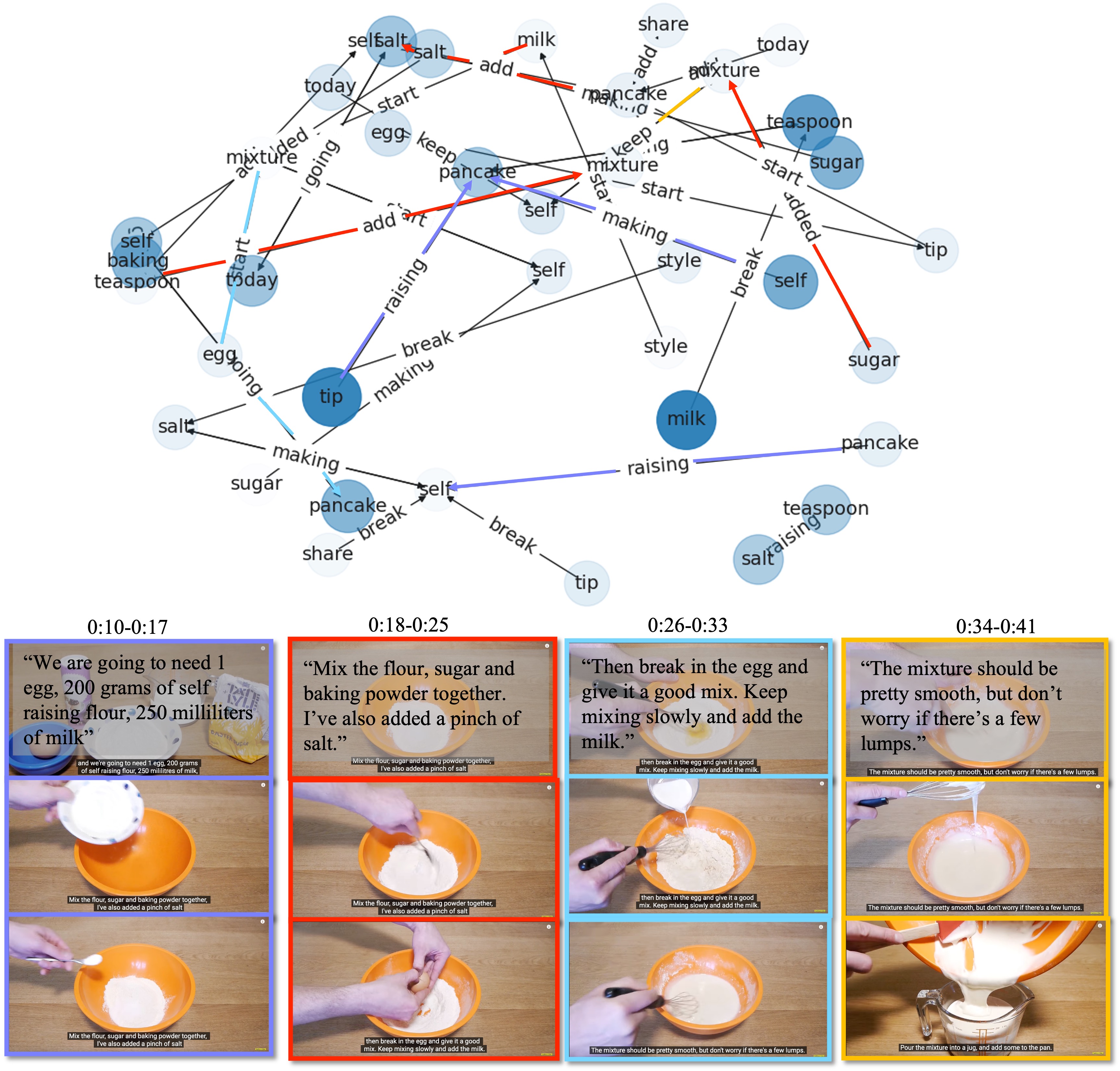}
    \caption{A larger example of ``How to make a pancake: Tips and Tricks" and its interpreted directed graph using SVGraph. The bottom row of images are sample frames from the video and associated narration. The instructions occurring at those times are highlighted by the same color to help guide understanding in this example.}
    \label{fig:making_pancakes}
\end{figure*}

\section{Additional Results}
\label{supp_additional_results}
Here we show some of the graphs in a larger format for better visibility. Figure \ref{fig:Change_Shower_head} and \ref{fig:making_pancakes} have sample frames from the original video for reference. For example, Figure \ref{fig:Change_Shower_head} shows an enlarged example from a video on ``How to Change a Shower Head". The graph shows many triplet relationships that are most relevant to the activity. For example, there is ``scratch$\rightarrow$ may $\rightarrow$ shower" which is often mentioned as a risk if a cloth is not used when unscrewing the old shower head with a plier. The most common activity is using pliers to remove the old shower head and adding the new shower head which as expected, is repeated throughout the graph. The graph shown in Figure \ref{fig:making_pancakes} is an additional example. In this example the addition of ingredients and mixing those ingredients appears to be the most common interactions in the graph. This shows the graph has highlighted the most important aspects of the instruction ``how to make pancakes". 

\begin{figure*}[t!]
    \centering
    \includegraphics[width=.85\linewidth]{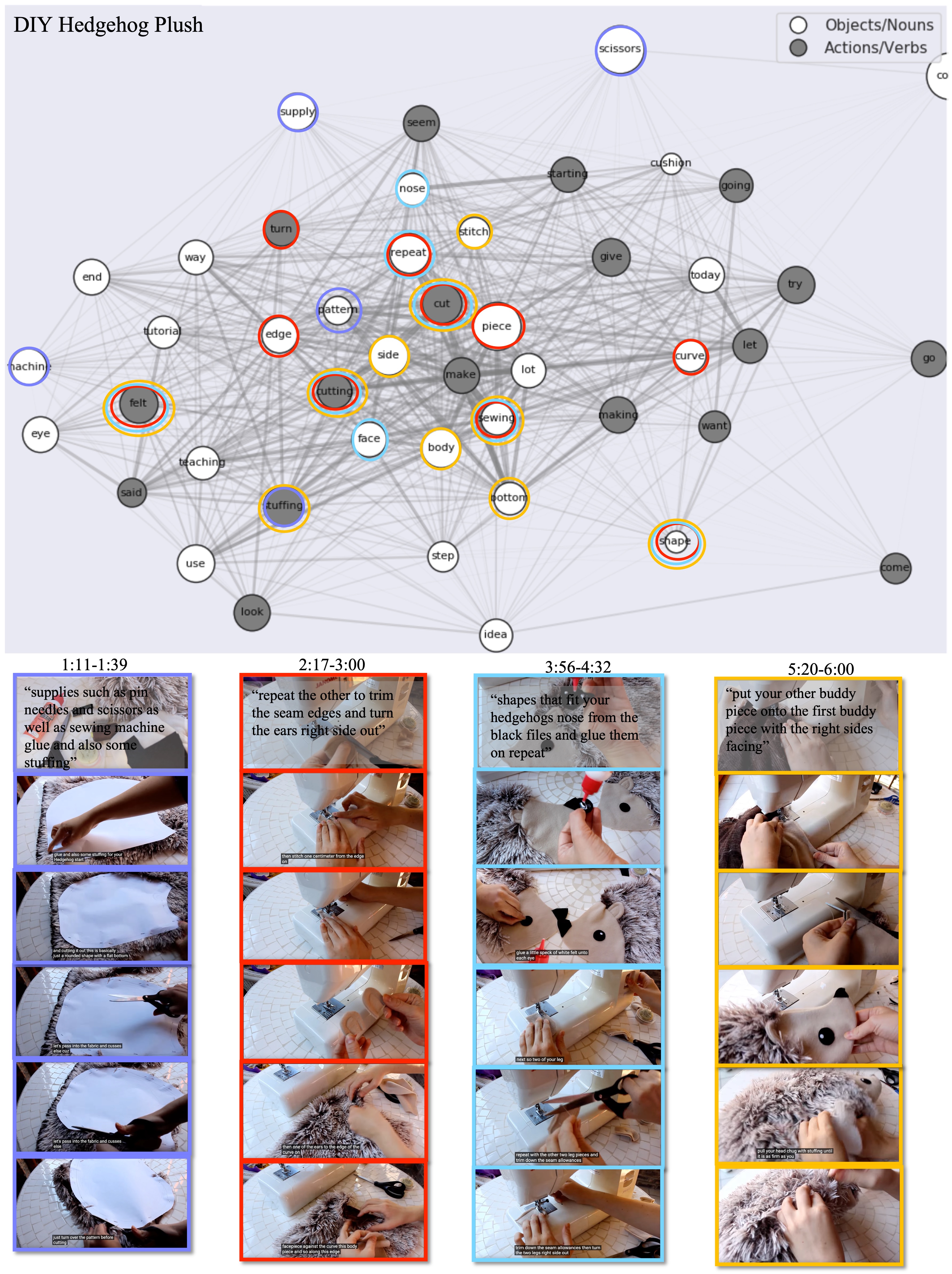}
    \caption{A larger example ``DIY Hedge Hog" and its interpreted undirected graph using SVGraph. The bottom row of images are sample frames from the video and associated narration. The instructions occurring at those times are highlighted by the same color to help guide understanding in this example.}
    \label{fig:diy_Hedge_hog}
\end{figure*}

Figure \ref{fig:diy_Hedge_hog} shows an enlarged example of an extracted undirected semantic graph where nouns and verbs are identified and labeled. The larger the node size, the greater the importance for the overall activity. The thicker the edges, the stronger the relationships between two nodes are overall. In this figure we highlight some of the concepts that match the sample frames for easier reference, however many concepts like ``sewing" and ``pattern" that are not highlighted are present in the instructional activity throughout the duration of the video. We see the important nodes are central to the graph and involve repeated activities such as “cut", “cutting", “sewing". Less common activities are more on the perimeter of the graph such as “end", “said", “come". This can also be seen in the objects/nouns.

\begin{figure*}
    \centering
    \includegraphics[width=.99\linewidth]{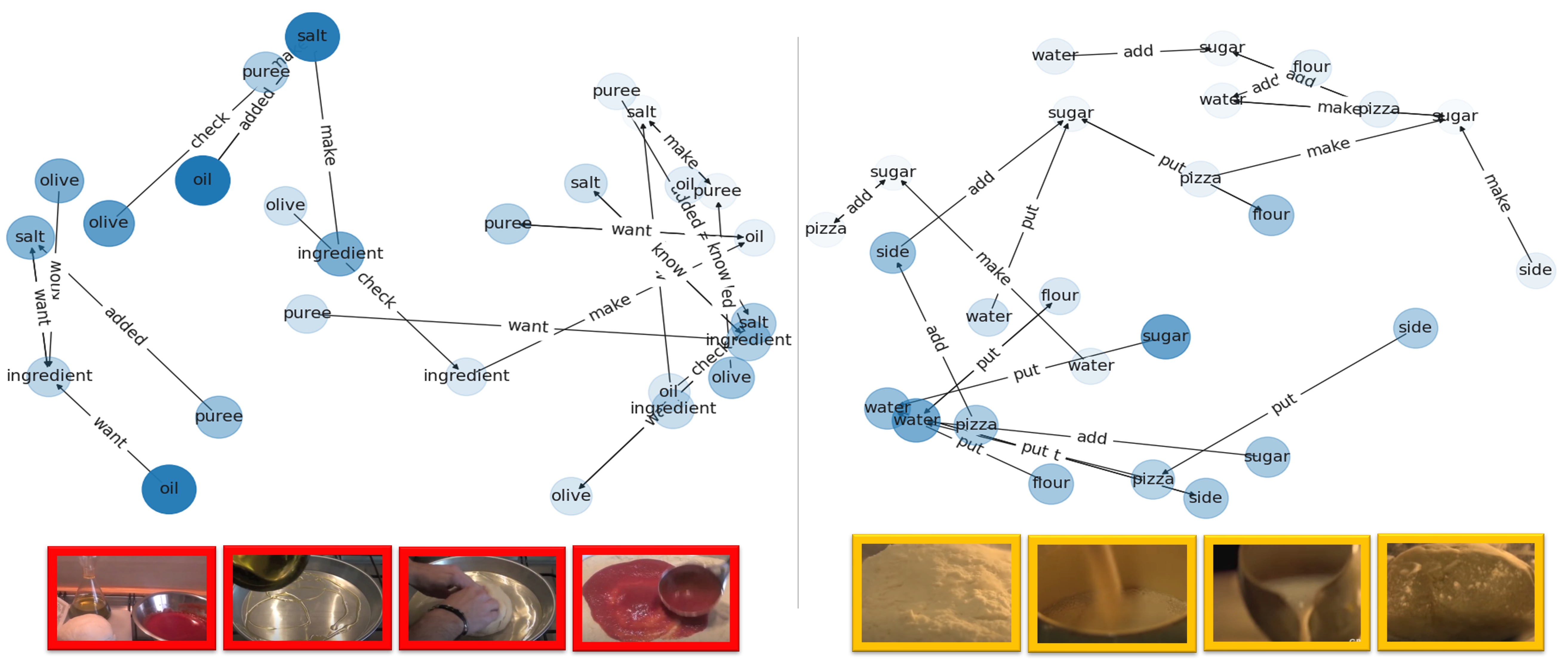}
    \caption{An enlarged example of two semantically interpreted graphs from instructional videos on ``How to Make a Margarita Pizza". }
    \label{fig:pizza}
\end{figure*}

\begin{figure*}
    \centering
    \includegraphics[width=.99\linewidth]{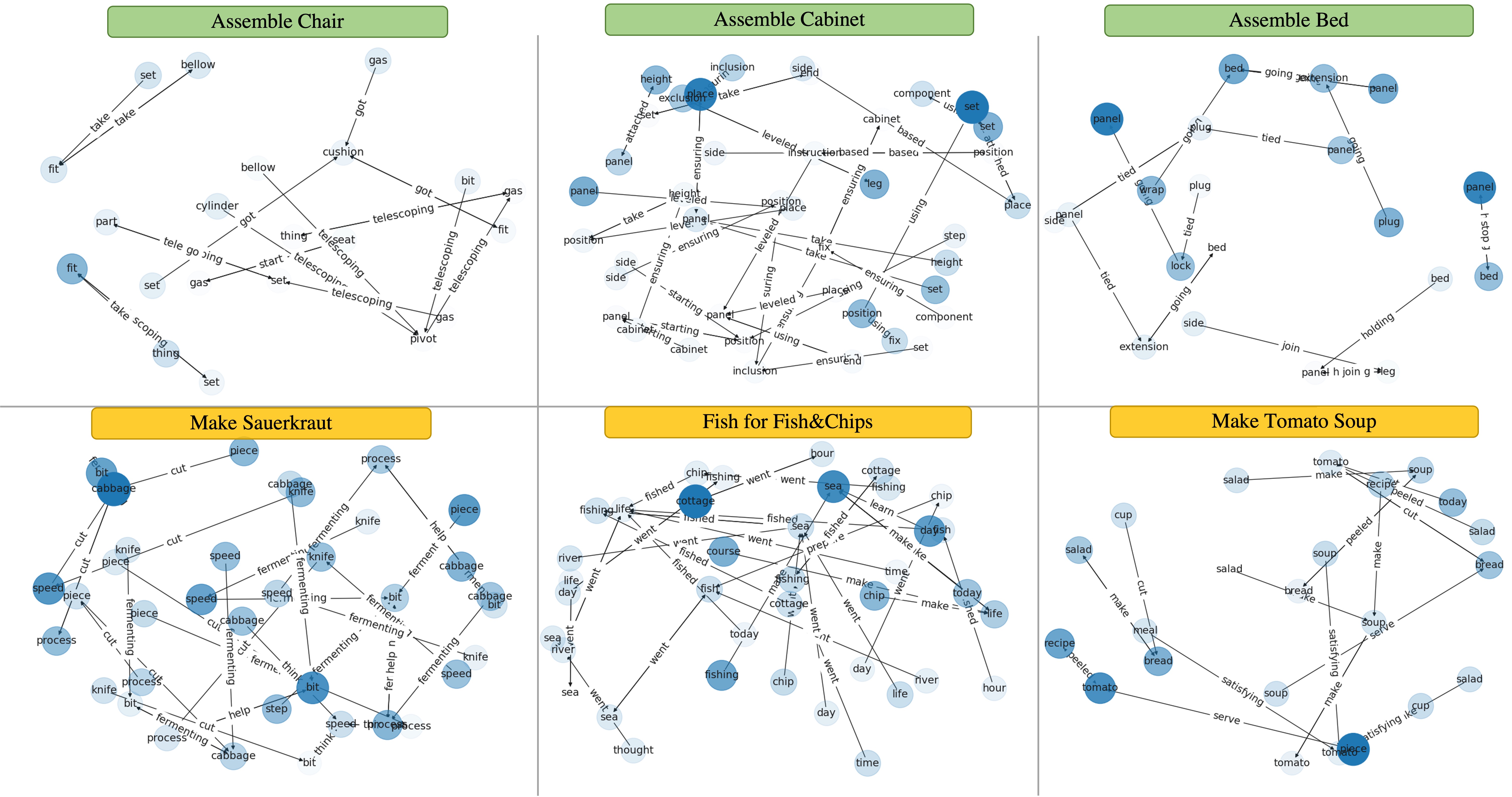}
    \caption{Additional samples of semantically interpreted graphs for a variety of tasks. }
    \label{fig:survey}
\end{figure*}

Figure \ref{fig:pizza} shows an enlarged graph comparison between videos showing the same instructional activity. We can see that while they are both categorized as making margarita pizza, they differ in many ways. The right focuses more on the dough aspect of making the pizza while the left focuses more on the added ingredients. This shows that our semantically interpreted graphs are more local, specific to the current video, rather than global, meaning learned representations across the same activity. Figure \ref{fig:survey} shows additional graphs from a variety of different tasks demonstrated in instructional videos. These graphs are very distinct, further supporting the localization of the interpretation of nodes and relationships to the particular instructional video and task. 

\end{document}